\documentclass[runningheads,a4paper]{llncs}
\usepackage{color}
\usepackage{subdepth}
\usepackage{amssymb}
\usepackage{graphicx}
\usepackage{epstopdf}
\usepackage{subfigure}
\usepackage{url}
\usepackage{amsmath}
\usepackage{multirow}
\usepackage{booktabs}

\long\def\/*#1*/{}

\begin{document}
\mainmatter  
\title{A Deep Learning based Joint Segmentation and Classification Framework for Glaucoma Assesment in Retinal Color Fundus Images}
\titlerunning{Joint Segmentation and Classification framework for Glaucoma assesment}
\author{Arunava Chakravarty \and Jayanthi Sivaswamy}
\tocauthor{Arunava Chakravarty, and Jayanthi Sivaswamy}
\authorrunning{A. Chakravarty et al.} % abbreviated author list (for running head)

\institute{Center for Visual Information Technology, IIIT Hyderabad, India,\\
\email{arunava.chakravarty@research.iiit.ac.in, jsivaswamy@iiit.ac.in}}
\maketitle

\begin{abstract}
Automated Computer Aided diagnostic tools can be used for the early detection of glaucoma to prevent irreversible vision loss. In this work we present a Multi-task Convolutional Neural Network (CNN) that jointly segments the Optic Disc (OD), Optic Cup (OC) and predicts the presence of glaucoma in color fundus images. The CNN utilizes a combination of image appearance features and structural features obtained from the OD-OC segmentation to obtain a robust prediction. The use of fewer network parameters and the sharing of the CNN features for multiple related tasks ensures the good generalizability of the architecture, allowing it to be trained on small training sets.  The cross-testing performance of the proposed method on an independent validation set acquired using a different camera and image resolution was found to be good with an average dice score of 0.92 for OD, 0.84 for OC and and AUC of 0.95 on the task of glaucoma classificationillustrating its potential as a mass screening tool for the early detection of  glaucoma.
\end{abstract}

\section{Introduction}
Glaucoma is a chronic retinal disorder which is a leading cause of blindness in the world. It is asymptomatic in its early stages and leads to a gradual but irreversible loss of sight. Large scale screening programs can help prevent its progression through early detection and treatment. Glaucoma is primarily characterized by the structural changes in the Optic Disc (OD) which appears as a bright elliptical structure in the color fundus images (CFI). It consists of a central depression called the Optic Cup (OC) which is surrounded by a concentric neuro-retinal rim. Glaucoma is characterized by the loss of retinal nerve fibers which results in the enlargement of the OC. The OC boundary is primarily characterized by the relative drop in depth from the OD boundary  \cite{chakravarty2014coupled}, \cite{chakravarty2017joint}. However, CFIs only provide a 2D view of the retinal surface and lack depth information. Hence, an accurate segmentation of OC in CFI remains an open problem.

The existing automated methods for glaucoma detection either rely on the explicit segmentation of the OD and OC to derive structural features \cite{sivaswamy2015comprehensive} and clinical parameters such as the Vertical Cup-to-disc ratio (CDR)\cite{cheng2013superpixel} to detect glaucoma, or employ  low-level appearance based image features \cite{maheshwari2017automated} to  directly classify the CFI into normal or glaucomatous categories. Alternatively, attempts at CDR estimation have also been made based on the reconstruction coefficients from a sparse representation \cite{cheng2017similarity} without the explicit segmentation of OD and OC. Few methods \cite{chakrabarty2016automated}, \cite{chakravarty2016glaucoma} have also attempted to combine both appearance and structural features to improve the classification performance. Recently, the Deep Learning based methods have led to a significant improvement in the state of the art. In \cite{fu2018joint}, a novel M-net architecture was explored to jointly segment the OD and OC in the log-polar domain and the CDR was extracted for glaucoma detection. The DENet architecture in \cite{fu2018disc} employed an ensemble of four independent streams of neural networks whose predictions were fused to obtain the final decision.

In this work we explore a single Multi-task deep Convolutional neural network (CNN) architecture which jointly performs the three tasks of the segmentation of OD, OC and image level prediction of glaucoma.   Sharing the CNN features for multiple but related tasks improves the generalizability of the network by constraining it to learn meaningful features. Moreover, the proposed method achieves a performance comparable to the state of the art with a relatively small network architecture that employs far fewer network parameters in comparison to the existing methods such as DENet. The sharing of the CNN features for multiple tasks and the smaller network size ensures that the proposed method can be trained in the limited availability of data without over-fitting. Our method also employs a combination of appearance as well as structural features obtained from the OD-OC segmentation within the CNN framework to improve the robustness of the glaucoma classification. 

\section{Method}

  \begin{figure}
 \centering
  \includegraphics[scale=.45]{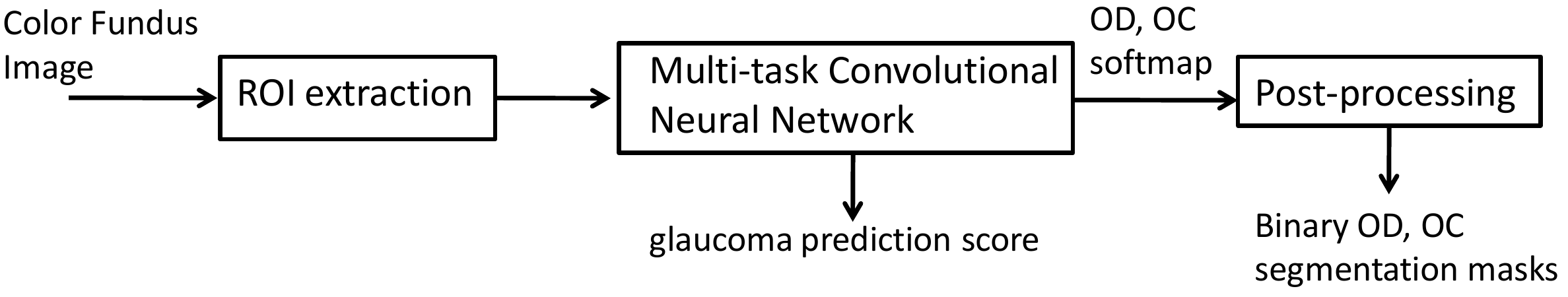}
 \caption{Outline of the proposed method.}
\label{block_diagram} 
 \end{figure}

An outline of the proposed method is depicted in Fig. \ref{block_diagram}. Given an input color fundus image, at first a rough Region of Interest (ROI) around the OD is extracted by employing a simple image processing based method described in \cite{chakravarty2017joint} based  on intensity thresholding and Hough Transform. Restricting further processing to the extracted ROI aids in decreasing the computational and memory requirements. The proposed method is relatively insensitive to the accuracy of the localization as long as the entire OD lies within the ROI, ie., the OD need not be present near the center of the ROI. Next, the extracted ROI is provided as an input to the proposed Convolutional Neural Network (CNN) architecture which jointly segments the OD, OC as well as provides an image level prediction for glaucoma. Further details of the CNN architecture is discussed in Section\ref{CNN}. The segmentation masks obtained from the CNN are further refined in a post-processing step (see Section \ref{postprocess}) to further improve their accuracy.

\subsection{The CNN architecture}
\label{CNN}

 \begin{figure}
 \centering
  \includegraphics[scale=.43]{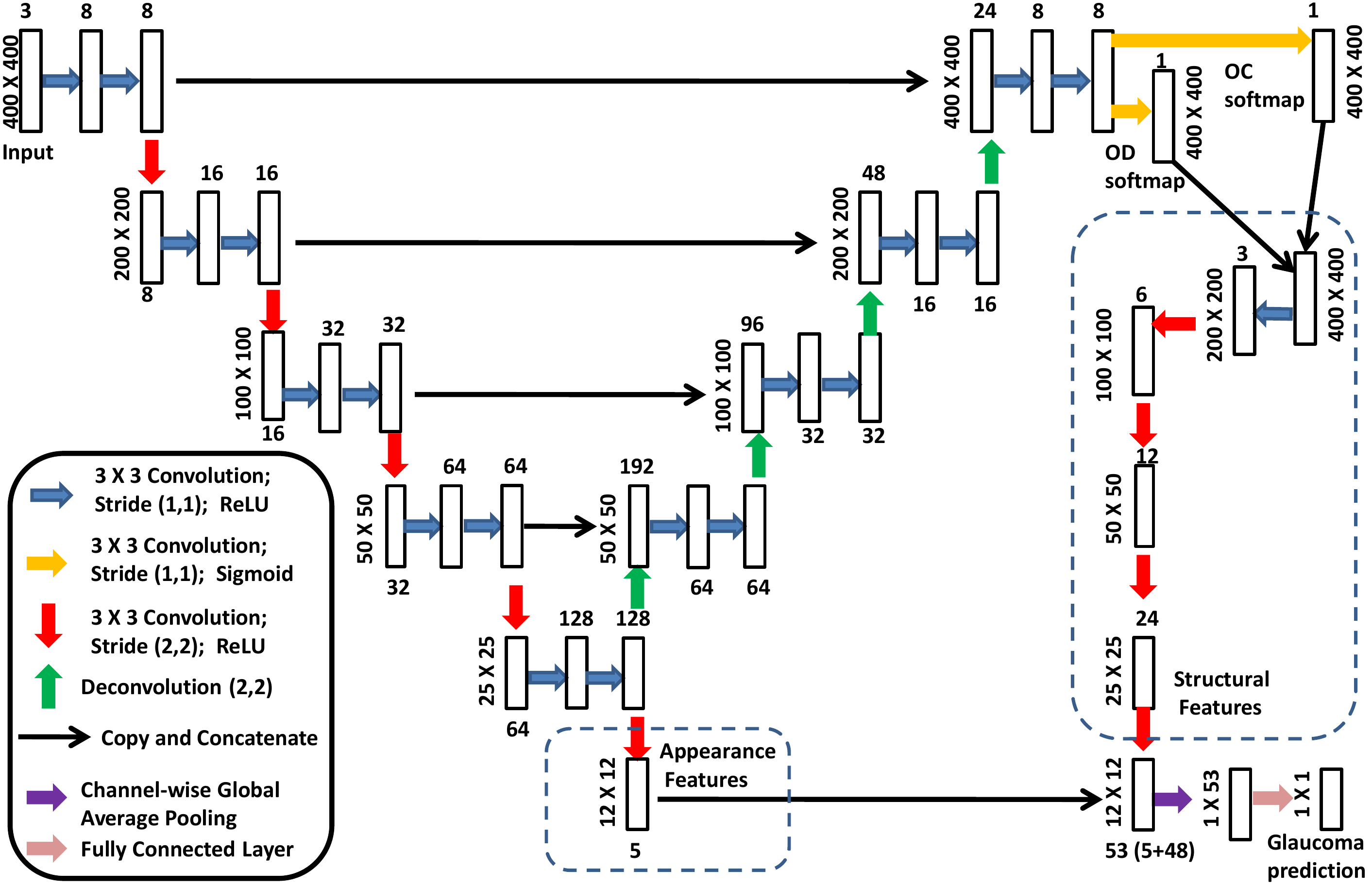}
 \caption{The proposed Multi-task Convolutional Neural Network architecture for joint segmentation of Optic disc, cup and image level classification of glaucoma using a combination of image appearance and structural features.}
\label{CNN_diagram} 
 \end{figure}

Fig. \ref{CNN_diagram} depicts the architecture of the proposed Multi-task CNN network. The input ROI image is resized to $400 \times 400$ and fed into a Deconvolutional network similar to the U-net \cite{ronneberger2015u}. It comprises an encoder network which successively downsamples the feature channels followed by a decoder network which successively upsamples to restore the original image resolution. Additional skip connections are provided between the corresponding encoder and decoder layers. The final output of the decoder network is fed into two separate  $3 \times 3$ convolution layers with a sigmoid activation function to obtain the output segmentation masks for the OD and OC. The classification part of the network uses a combination of image appearance and structural features (obtained from the OD-OC segmentation). The appearance features are obtained by reusing the $25 \times 25 \times 128$ output of the encoder part of the U-net architecture and applying a $5$ filter convolutional layer of size $3 \times 3$ and stride $(2,2)$ to obtain a $12 \times 12 \times 5$ size feature. In order to obtain the structural features, the output OD and OC segmentation masks are concatenated to obtain a 2 channel feature map and a series of $3 \times 3$ convolutional layers with a stride of (2,2) is successively applied to obtain the high-level structural features of size $12 \times 12 \times 48$. Finally, the appearance (5 channels) and the structural features (48 channels) are concatenated and a channel-wise Global average pooling is applied to obtain a 53-D feature vector. A fully connected layer with a single neuron is applied to the feature with a sigmoid function to obtain the probability of the presence of glaucoma.

\subsection{Implementation details}
A loss function has to be defined for each of the three outputs of the multi-task CNN network. The binary cross-entropy was used as the loss function for the image level classification task. The dice coefficient was used as a measure of the extent of overlap between the ground truth (GT) and the predicted segmentation masks. $L_{od}=1-Dice(R_{od},Y_{od})$ and $L_{cp}=1-Dice(R_{cp},Y_{cp})$ were used as the loss functions for the OD and OC segmentation tasks respectively, where $R_{od}$ and $R_{cp}$ denotes the segmentation masks computed by the network and $Y_{od}$, $Y_{cp}$ represents the GT segmentation masks for the OD and OC respectively. The Dice metric was defined similar to \cite{chakravarty2018race} as $Dice(R,Y)=\frac{2 \sum _{i} r_{i}. y_{i}}{\left( \sum _{i} r_i+\sum_{i} y_{i}\right)} $, where $r_i$ and $y_{i}$ represents the value at the $i^{th}$ pixel in $R$ and $Y$ respectively and the summations run over all the pixels in the image. The Dice metric is differentiable (see \cite{chakravarty2018race} for further details). The proposed CNN architecture is trained using gradient backpropagation with the ADAM algorithm \cite{kingma2014adam} to automatically adapt the learning rate. The batch size was fixed to 32 images and early stopping was used to terminate the training when the validation loss didnot improve for 20 consecutive epochs. The Data Augmentation plays an important role in the proper training of the CNN. Simple geometric transformations such as translation and rotation were incorporated to ensure that the proposed method is not sensitive to slight changes in the view and localization errors incurred during the ROI extraction step. The REFUGE dataset also offers additional challenges. First,the training dataset has a huge class imbalance with a 1:9 ratio between the number of glaucomatous and normal images. This approximates the expected distribution of images in a mass screening scenario where majority of the subjects are Normal. The second challenge is the large diversity between the training and test set. They were acquired at different resolutions, using different fundus cameras and a large difference in the distribution of the color intensity values in the RGB channels can be observed across the two dataset. In order to address these issues, the intensity of the training images were also modified during the data augmentation to ensure that the proposed system is invariant to changes in the color distribution. The RGB channels of the training images were transformed using the method in \cite{xiao2006color} such that the mean and standard deviation of the color channels were mapped to randomly sampled values from a gaussian distribution. Moreover, a PCA based technique similar to \cite{krizhevsky2012imagenet} was also employed to add some random noise to the images. The class imbalance was handled by augmenting a larger number of glaucomatous images to the training set in comparison to the Normal images.

\subsection{Post-processing}
\label{postprocess}
Since, the CNN poses the OD and OC segmentation as binary pixel-labeling problems, it doesnot explicitly constrain the segmented regions to be a single connected component or preserve the smoothness constraints on the OD and OC boundaries. In order, to address these issues, a post-processing step is employed to further refine the segmentation results obtained from the CNN. During post-processing, both the OD and OC segmentation softmaps are binarized by thresholding at 0.5. Thereafter, a morphological opening operation is employed to remove spurious small regions and smooth the boundaries. Finally, a connected component analysis is performed to remove all except the largest connected component in the binary segmentation mask. Though OC boundaries can have an arbitary shape, OD boundary can be closely approximated by an ellipse. Hence, an additional step is employed for the OD segmentation where an ellipse is fitted to it with the minimum least square error. Note, that the segmentation feature based pathway within the CNN which is used to predict the presence of glaucoma employs the original segmentation output of the CNN without the post-processing.

\section{Result}
\textbf{Dataset: } The performance of the proposed method has been evaluated on the REFUGE dataset. The training dataset consists of 400 (360 Normal + 40 Glaucomatous cases) Zeiss Visucam images of resolution $2124\times2056$ pixels. The image level diagnosis for glaucoma and the manual annotations of the Optic Disc and Cup are provided as the GT for each image. An off-site validation set of 400 Canon CR-2 images ($1634\times1634$ pixels) is also provided to evaluate the performance of the participating methods. \textit{The Ground truth of these images are not publicly available and the validation performance of the submitted results are evaluated by the challenge organizers. }

\textbf{Experimental Setup: } A four-fold cross-validation was performed on the training dataset by randomly dividing it into four parts of 100 images (90 Normal+10 Glaucomatous) each. The performance of the proposed method was evaluated on each part after training it on the remaining images. Thus, four CNNs with identical architecture but different network weights were learned (one for each fold). During testing on the off-site validation set, four predictions for each image were obtained using each of the four models learned during the cross-validation and their average was computed to obtain the final prediction.

\begin{figure}
 \centering
  \includegraphics[scale=.3]{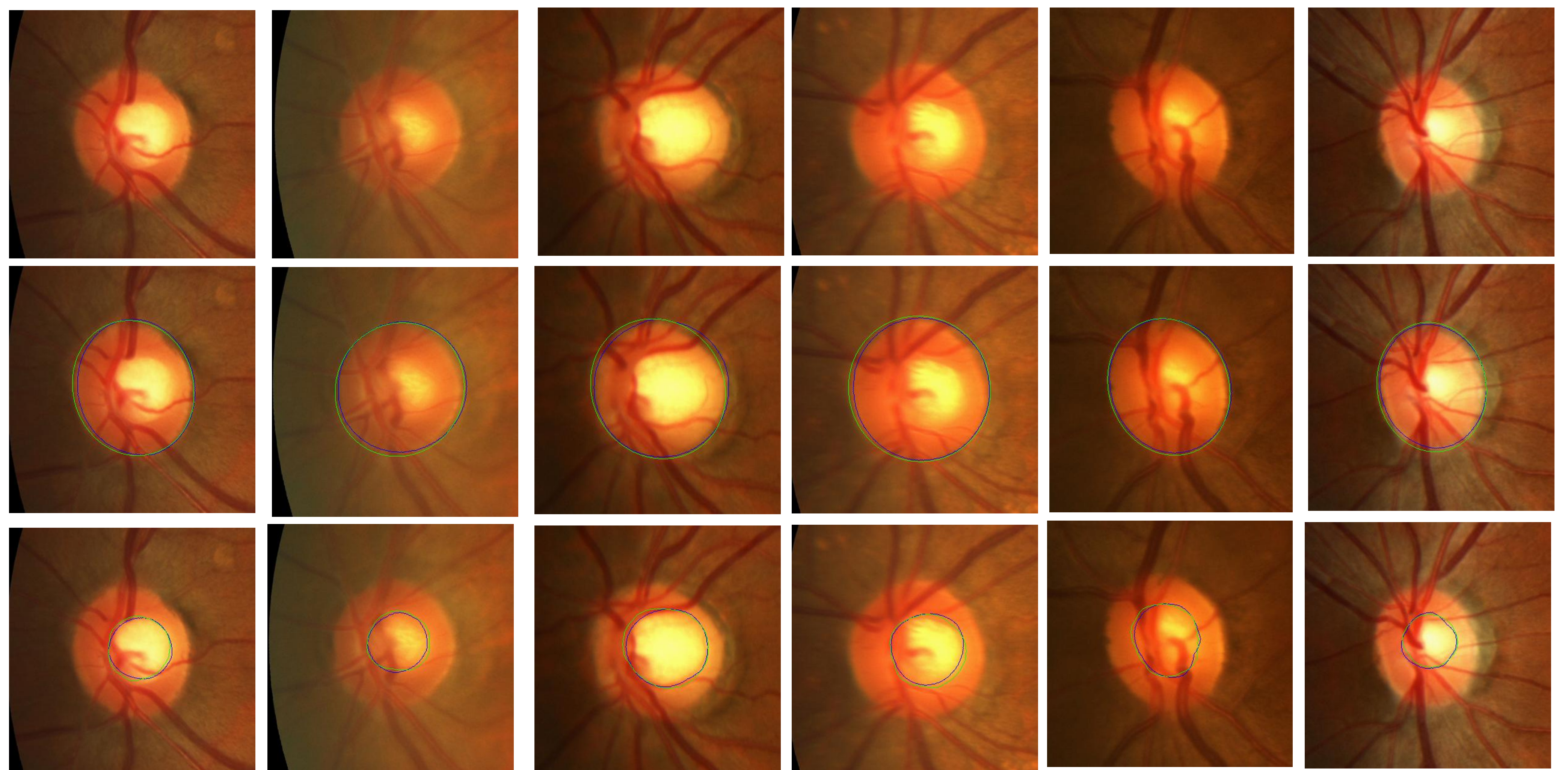}
 \caption{Cross-validation results of Optic Disc and Cup Segmentation: $1^{st}$ row depicts the ROI; OD and OC boundaries are presented in the $2^{nd}$ and $3^{rd}$ rows respectively. The proposed method's output (in green) is compared to the GT markings (in blue).}
\label{qual_res} 
 \end{figure}

\textbf{Performance of Optic Disc and Cup Segmentation: } The segmentation performance for  the four-fold cross-validation on the training set as well as the validation set is depicted in Table \ref{Segmentation} and sample qualitative results are depicted in Fig. \ref{qual_res}.

\renewcommand{\tabcolsep}{12pt}
\begin{table}[]
\centering
\caption{Performance of Optic Disc and Cup Segmentation. (mean $\pm$ standard dev.)}
\label{Segmentation}
\resizebox{\columnwidth}{!}{
\begin{tabular}{@{}lccc@{}}
\toprule
                           & Optic Disc Dice & Optic Cup Dice & Mean CDR error (px) \\  \midrule
Cross-validation  &       0.96 $\pm$ 0.07  & 0.86 $\pm$ 0.09  & 0.05 $\pm$ 0.09                    \\ 
Off-site Validation    &      0.92           &       0.84          &     0.05                \\  \bottomrule
\end{tabular}
}
\end{table}

\begin{figure}
 \centering
  \includegraphics[scale=.5]{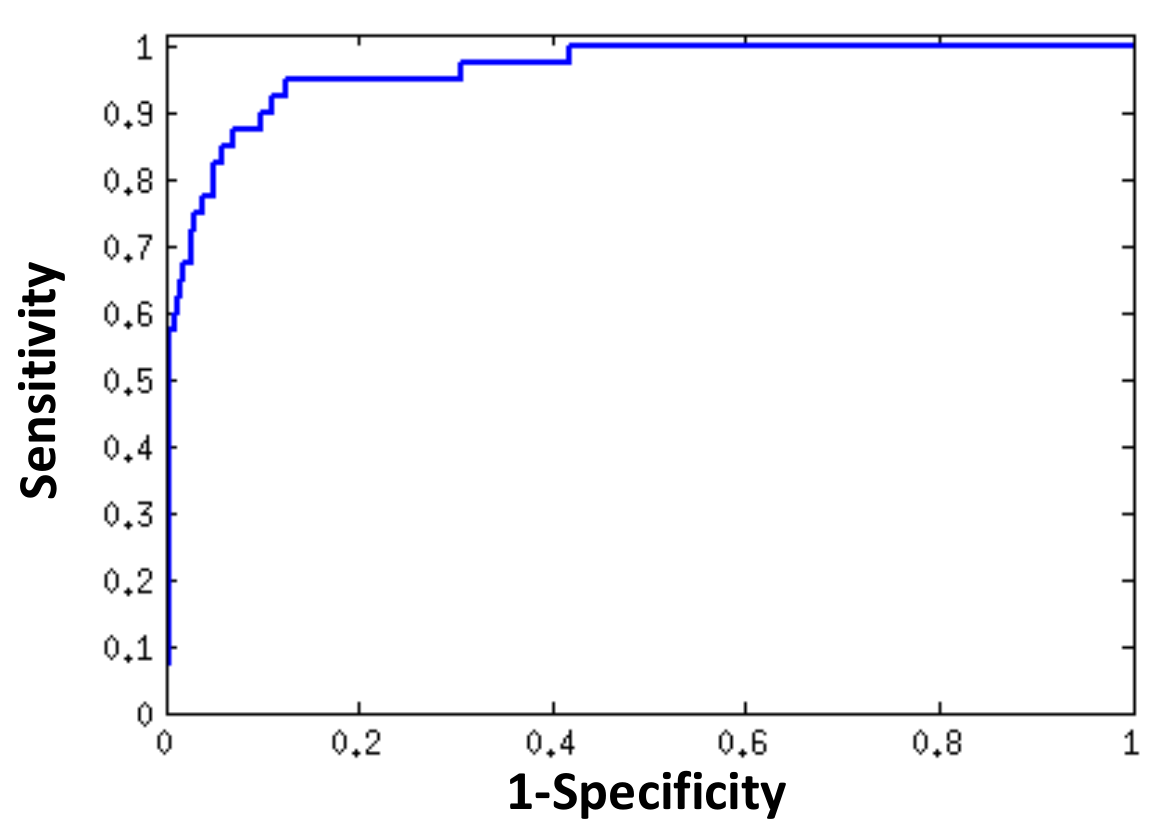}
 \caption{Average cross-validation Area Under the ROC curve on the training set.}
\label{res} 
 \end{figure}

\textbf{Classification Performance :} The classification performance of the proposed method is provided in Table \ref{classification} and the  ROC plot of the four fold cross-validation is presented in Fig. \ref{res} respectively. The sensitivity and specificity for the cross-validation experiment is reported at the optimal cutoff point of 0.575 which maximizes the Youden's index (Sensitivity+ Specifity-1). Though the training and the validation set images are acquired using different cameras and at different resolutions, the results donot show any significant difference in performance between the two datasets indicating the robustness of the method.

\renewcommand{\tabcolsep}{12pt}
\begin{table}[]
\centering
\caption{Performance of Glaucoma Classification}
\label{classification}
\resizebox{.7 \columnwidth}{!}{
\begin{tabular}{@{}lccc@{}}
\toprule
                   & Sensitivity & Specificity & AUC \\ \midrule
Cross-validation   &   0.88          &    0.91         &  0.96   \\ 
Off-site Validation &   ---          &     ---        &  0.95   \\ \bottomrule
\end{tabular}
}
\end{table}

\vspace{-10pt}

\vspace{-10pt}
\section{Discussion and Conclusion}
In this work we have explored a novel Multi-task CNN architecture to jointly segment the OD, OC and predict the presence of glaucoma in CFI. The features of the CNN are shared across all the three related tasks thereby reducing the computational requirements and improving the generalizability of the learned features. The proposed architecture employs $609,170$ network parameters and is significantly smaller in comparison to the existing state of the art architectures such as DENet \cite{fu2018disc}. Fewer network parameters ensure that the network doesnot overfit when trained on small datasets. Though the images in training and test set are acquired using different fundus cameras and at different resolutions, the proposed method didnot show a large drop between the cross-validation and cross-testing performance indicating the robustness of the method. The performance of the proposed method is comparable to the state of the art with a dice coefficient of 0.92 for OD segmentation, 0.84 for OC segmentation and an AUC of 0.95 on the task of glaucoma classification on the test set illustrating its potential as a mass screening tool for the early detection of  glaucoma.

\section*{Acknowledgement}
We would like to thank the organizersof the Retinal Fundus Glaucoma Challenge (REFUGE) ($https://refuge.grand-challenge.org/$) for kindly providing the dataset and evaluating the performance on the off-site validation set that has greatly assisted this work.

\bibliographystyle{splncs03}
\bibliography{ref_short1}

\end{document}